\documentclass[11pt]{article}
\usepackage[margin=1in]{geometry}
\usepackage{amsmath}
\usepackage{amssymb}
\usepackage{graphicx}
\usepackage{booktabs}
\usepackage{longtable}
\usepackage{array}
\usepackage[dvipsnames]{xcolor}
\usepackage[font=small,labelfont=bf]{caption}
\usepackage[numbers,sort&compress]{natbib}
\usepackage[colorlinks=true,allcolors=NavyBlue,breaklinks=true]{hyperref}

\title{\textbf{Cheap Verifiers, Large Blind Spots:\\[2pt]
Measuring the Reliability Cost of Cost-Saving Cascades}}
\author{Dushyant Rajput \quad Nirdesh Chauhan \quad Siddharth Kosaraju\\[4pt]
{\normalsize AltSlate Labs LLP}\\[2pt]
{\small\texttt{dushyant@altslate.com} \quad \texttt{nirdesh@altslate.com} \quad \texttt{siddharth@altslate.com}}}
\date{August 31, 2026}

\begin{document}
\maketitle

\begin{abstract}
Inference cascades cut cost by answering most queries with a cheap model and
escalating a hard tail to a frontier model that acts as verifier. A natural
extension closes the loop---fine-tune the cheap student on the verifier's
rejections so the escalation rate, and cost, fall each round. We set out to
measure this loop on real LLMs, and report four findings. First, the verifier's
\emph{blind spot}---the fraction of the student's wrong answers it waves
through---is large and moves adversarially: it \emph{grows} with student
capability ($\beta$ from $0.12$ to $0.55$ as the student scales 0.5B$\to$32B) and
\emph{shrinks} with verifier capability, so it is worst exactly in the
cheap-student, cheap-verifier configuration cascades exist to create. Second,
buying it away returns the saving: a frontier verifier drives $\beta$ to
${\approx}\,0.05$ but then escalates on $46\%$ of hard-MATH queries against a
$39\%$ true error rate---paying the frontier price on nearly half of all traffic,
the very cost the cascade exists to avoid. Third, naive corrective fine-tuning on
the verifier-rejected tail does not improve the small student but \emph{degrades
and ultimately collapses} it, across every teacher we tried (cross-family and
same-family)---so at this scale the ``self-improving'' loop is self-defeating.
Fourth, throughout all of this the cascade's own dashboard---every metric computed
through the verifier---reads a flat ${\approx}\,3\%$ error while true delivered
error swings up to $32\%$: the system is blind to its own degradation \emph{by
construction}. We then give the theory that explains the blindness---a
two-population \emph{conservation law}, $\varepsilon_\infty \lesssim q_0\beta_0$,
under which every in-loop metric improves while true quality does not---and a
synthetic study that validates the mechanism where the blind spot's dynamics are
emergent rather than imposed. The practical conclusion is a measurement
discipline: the reliability of a self-improving cascade cannot be read from any
metric computed through its own verifier.
\end{abstract}

\vspace{1em}

\section{Introduction}

The dominant lever for reducing the cost of large-language-model
inference has shifted from the model to the \emph{harness} --- the
scaffold of routing, verification, retrieval, and retries wrapped around
a fixed set of weights. Among harness techniques, the \emph{cascade} is
the most direct cost play: run a cheap model on every query, and
escalate to an expensive frontier model only when some signal says the
cheap answer is untrustworthy
\cite{chen2023frugalgpt,dohan2022cascades,aggarwal2023automix}.
The frontier model functions as a verifier; the cascade pays its price
only on the escalated tail.

A tempting extension closes the loop. Every escalation yields a
frontier-quality answer on exactly the distribution where the cheap
model fails --- a free training example. Fine-tune the cheap
\emph{student} on these corrections, and its error rate falls, its
escalation rate falls with it, and the cascade gets cheaper each round.
Recent work explores online, training-free versions of this idea, in
which deferred queries produce reusable in-context strategies for the
weak model \cite{intercascade2025,sarukkai2025deferral}; the
parametric version, which updates the student's weights, is the natural
and more aggressive sibling.

The loop invites an analogy to speculative decoding, where a small
drafter proposes tokens that a large model verifies in parallel, cheaply
and --- crucially --- \emph{losslessly}: rejection sampling guarantees
the output distribution is exactly the large model's
\cite{leviathan2023speculative,chen2023speculative}. If semantic
cascades inherited that guarantee, the self-improving loop would be a
strict win: lower cost, preserved quality. They do not. Speculative
decoding verifies a \emph{token} against a distribution it has in closed
form; a semantic cascade verifies a \emph{claim} against a judgment the
verifier must itself infer, and that judgment is wrong in both
directions. Once verification is imperfect, training a student to
satisfy the verifier is not distillation toward truth --- it is
optimization against a proxy, the setting in which reward-model
overoptimization is known to arise
\cite{gao2023scaling,skalse2022reward}, a modern instance of
Goodhart's law \cite{manheim2018goodhart}.

This paper leads with what we \emph{measured}. We built the wind tunnel
needed to see a verifier's blind spot --- tasks with a cheap oracle from
which the verifier is deliberately blinded (Section~\ref{sec:design}) ---
and ran the loop on real language models (Qwen2.5 students 0.5B--32B,
GSM8K and hard MATH, OpenAI verifiers up to gpt-5-mini). Four findings,
all measured rather than assumed, organize the paper.

\begin{itemize}
\item
  \emph{The blind spot is worst exactly where cascades operate.} The
  verifier's blind-spot rate \(\beta\) --- the fraction of the student's
  wrong answers it accepts --- \emph{grows} with student capability
  (from \(0.12\) at 0.5B to \(0.55\) at 14B, fixed verifier) and
  \emph{shrinks} with verifier capability. A more capable student makes
  subtler, more convincing errors; a cheaper verifier catches fewer of
  them. The danger zone is therefore the cheap-student, cheap-verifier
  corner that makes cascades attractive in the first place
  (Section~\ref{sec:real}).
\item
  \emph{Buying the blind spot away gives the cost back.} Swapping a
  frontier verifier in on hard MATH collapses \(\beta\) to
  \(\approx 0.05\), but that verifier then escalates on \(46\%\) of
  queries against a \(39\%\) true error rate --- it buys recall by
  over-rejecting, paying the frontier price on nearly half of all
  traffic. Low blind spot and low cost are not simultaneously available
  from a fixed verifier.
\item
  \emph{The ``self-improving'' loop, at this scale, is self-defeating.}
  Naive corrective fine-tuning on the verifier-rejected tail did not
  improve any small student we tried; it \emph{degraded} them and,
  cumulatively, \emph{collapsed} them --- across cross-family and
  same-family teachers alike. We never instantiated a loop that improved
  the student, and report that plainly: it is a direct caution against
  the \emph{parametric} self-improving loop (fine-tuning the student on
  the verifier's rejects), the natural next step beyond today's
  training-free deferral-reuse methods, and it means the clean error
  \emph{floor} below stays a theoretical result rather than a measured
  one.
\item
  \emph{None of this is visible from inside.} Every metric a
  practitioner monitors is computed through the verifier, and reads a
  flat \(\approx 3\%\) error while true delivered error swings to
  \(32\%\) --- the dashboard cannot distinguish a healthy student from
  one the loop is actively wrecking.
\end{itemize}

The rest of the paper explains \emph{why} the dashboard is blind. We
formalize the loop and its one structural property --- training signal
derived \emph{only} from verifier-detectable errors
(Section~\ref{sec:loop}) --- and derive a \emph{conservation law}:
user-facing error asymptotes not to zero but to a floor anchored at the
initial confidently-wrong-and-accepted mass,
\(\varepsilon_{\infty} \lesssim q_{0}\beta_{0}\) (Section~\ref{sec:law},
with a two-population proof in Appendix~\ref{sec:appendix}), under which
every verifier-computed metric improves while true quality does not. A
synthetic mechanism study, where the blind spot's dynamics are
\emph{emergent} rather than imposed, validates the law and the
mitigation exchange rates (Section~\ref{sec:exp}). The theory is the
explanation; the measurements are the result.

\section{Background and related work}

\textbf{Cascades and routing.} Cascades query models in sequence and
decide, from a post-generation signal, whether to accept the cheap
answer or escalate
\cite{chen2023frugalgpt,dohan2022cascades,aggarwal2023automix}.
Routers instead decide \emph{before} generation which model should
answer \cite{ong2024routellm}. Both aim at the cost--quality Pareto
frontier; both, in their standard form, hold the cheap model
\emph{fixed}. Our object of study is the case where the cheap model is
not fixed but is trained on the cascade's own escalation signal, which
changes the dynamics qualitatively.

\textbf{Verifier-as-training-signal.} Using a stronger model or a
checker to supervise a weaker generator is the backbone of
rejection-sampling fine-tuning and reinforced self-training
\cite{gulcehre2023rest}, self-taught reasoning \cite{zelikman2022star},
and weak-to-strong supervision studies \cite{burns2023weaktostrong}.
Recent cascade work makes the loop explicit and online, storing the
strong model's deferral-time strategies for reuse
\cite{intercascade2025,sarukkai2025deferral}, the latter deferring
by self-consistency \cite{wang2022selfconsistency}. This literature
reports accuracy-at-cost, typically as a single snapshot. None of it, to
our knowledge, characterizes what happens to the \emph{composition} of
the student's residual errors as the loop runs, which is precisely where
the blind-spot effect lives.

\textbf{Overoptimization and imperfect verifiers.} Optimizing a policy
against a learned proxy of human preference improves the proxy's score
while eventually degrading the true objective --- the gold reward turns
over even as the proxy reward climbs
\cite{gao2023scaling,skalse2022reward}, a modern Goodhart effect
\cite{manheim2018goodhart}. This is the phenomenon underneath our claim,
but the signature we predict is \emph{not} Gao et al.'s turnover: under
the corrective loop the proxy improves while gold error stays \emph{flat
at a positive floor}, and a turnover can appear only when accepted
outputs are recycled as labels (self-training, our H3). Closest to us,
Stroebl et al. \cite{stroebl2024inference} show that inference-time
\emph{resampling} against an imperfect verifier cannot beat the
verifier's false-positive rate --- a static, single-shot form of the
blind spot. Our claim is its closed-loop counterpart: once the student
is \emph{trained} against the verifier, that same false-accept mass
becomes a round-persistent error floor rather than a per-query bound ---
pushed down only by generalisation spillover (corrective) and held
higher when accepted outputs are recycled (self-training) --- and turns
invisible to every in-loop metric. Beyond the shape difference, our
setting differs from reward-model overoptimization in that the proxy is
the \emph{expensive resource whose invocation the loop is trying to
minimize} --- so the overoptimization dose and the cost saving are the
same axis, a coupling absent from RLHF --- and that the student is
trained on a \emph{self-selected slice} (only the verifier's rejects)
rather than a fixed preference distribution.

\textbf{LLM-as-judge and self-preference.} Using an LLM to judge another
model's output is now standard \cite{zheng2023judging}, as is the finding
that judges can favor text from their own family or their own
generations \cite{panickssery2024selfpreference}. We distinguish that
effect from ours. Our correlation corollary is not about a judge
preferring its own \emph{text} it is about shared \emph{failure modes}
--- a distractor whose reasoning ``looks right'' to a student built on
certain pretraining priors also looks right to a verifier built on
similar priors. Crucially, we make the causal claim on a \emph{measured}
blind-spot rate, not on a family label, so it does not depend on the
self-preference mechanism being the cause.

\textbf{Self-consuming loops and pseudo-labeling.} Training a model on
its own outputs can cause \emph{model collapse} --- loss of variance and
tails under recursive generation
\cite{shumailov2024collapse,alemohammad2024selfconsuming}. Our
self-training loop is a \emph{verifier-filtered} self-consuming loop,
and the difference is the point: there, collapse arises from unfiltered
recursion; here, the verifier's blind spot \emph{selects} which
self-outputs are reinforced, concentrating error precisely where the
filter cannot see. At the level of one round this is the classic
\emph{confirmation bias} of pseudo-labeling \cite{arazo2020pseudo} and a
reason intrinsic self-correction stalls without external signal
\cite{huang2024selfcorrect}; what those results lack --- and what the
cascade supplies --- is that the filter is an \emph{external} frontier
verifier, so the floor is set by student--verifier blind-mass overlap
rather than by the student's own confidence.

\textbf{Selective prediction and learning to defer.} The verifier's
accept/reject decision is a deferral rule, and its blind-spot rate
\(\beta_{t}\) is the miscoverage of that rule --- the quantity selective
classification \cite{elyaniv2010selective} and learning-to-defer
\cite{mozannar2020defer} are built to control. That literature studies
the risk--coverage tradeoff of a \emph{static} rule; we study the
closed-loop dynamics that arise once the deferred slice becomes training
data and the rule's blind region is never corrected.

\section{The self-improving cascade}

\protect\phantomsection\label{sec:loop}{}

We fix notation. A \emph{student} \(S\) maps a query \(x\) to an output
\(S(x)\). A \emph{verifier} \(V\) maps a query and a candidate output to
a binary decision,
\(V(x,y) \in \left\{ \text{accept},\text{ reject} \right\}\). A
\emph{teacher} \(G\) (often the same frontier model as \(V\)) produces a
replacement output \(G(x)\) on rejection. An \emph{oracle}
\(O(x,y) \in \left\{ \text{correct},\text{ wrong} \right\}\) gives
ground truth; the oracle exists in analysis but is \emph{not available
to the loop} --- if it were, one would simply verify with it.

\begin{figure}
\centering
\includegraphics[width=0.78\linewidth,height=\textheight,keepaspectratio]{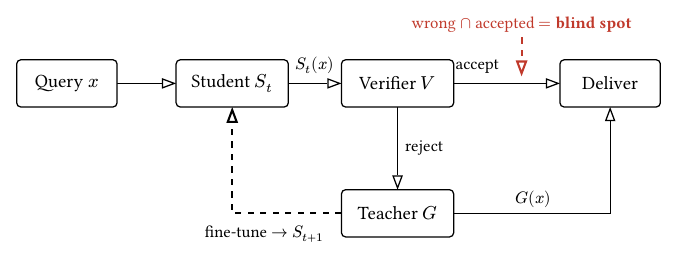}
\caption{The self-improving cascade. The student's output is either
accepted and shipped, or rejected and replaced by the teacher's; in the
corrective loop only rejected items become training signal (dashed),
while the self-training variant also feeds accepted outputs back as
labels. Errors the verifier accepts by mistake are shipped unflagged
and, in the corrective loop, never enter training --- the blind spot.}
\label{fig:loop}
\end{figure}

The cascade delivers \(S(x)\) when \(V\) accepts and \(G(x)\) when \(V\)
rejects (Figure~\ref{fig:loop}). The \emph{self-improving} variant
additionally updates the student. In round \(t\):

\begin{enumerate}
\item
  Draw a batch \(B_{t}\); the student \(S_{t}\) generates \(S_t(x)\)
  for \(x \in B_{t}\).
\item
  The verifier judges each output; let
  \(R_{t} = \left\{ x:V\left( x,S_t(x) \right) = \text{ reject} \right\}\)
  and \(A_{t} = B_{t} \smallsetminus R_{t}\).
\item
  Form training targets and fit \(S_{t + 1}\). The \emph{corrective}
  loop trains on \(\left\{ \left( x,G(x) \right):x \in R_{t} \right\}\).
  The \emph{self-training} loop adds the accepted student outputs
  \(\left\{ \left( x,S_t(x) \right):x \in A_{t} \right\}\) as positive
  targets.
\end{enumerate}

Two quantities drive everything. Let

\[q_{t} = \Pr\left\lbrack O\left( x,S_t(x) \right) = \text{ wrong} \right\rbrack\]

be the student's \emph{raw error rate}, and let

\[\beta_{t} = \Pr\left\lbrack V\left( x,S_t(x) \right) = \text{ accept }\,~|~\, O\left( x,S_t(x) \right) = \text{ wrong} \right\rbrack\]

be the verifier's \emph{blind-spot rate} --- the fraction of the
student's genuine errors that the verifier waves through. Its complement
\(r_{t} = 1 - \beta_{t}\) is the verifier's recall on errors.

The structural fact that organizes the rest of the paper is this:
\emph{the loop's training signal is a function of \(V\), never of
\(O\)}. Rejections identify errors \(V\) can see; the errors \(V\)
cannot see are, by construction, indistinguishable to the loop from
correct answers. The loop is therefore a selection process that removes
verifier-detectable errors and retains verifier-undetectable ones.

\section{Measuring a blind spot: the wind-tunnel method}

\protect\phantomsection\label{sec:design}{}

The obstacle to measuring any of this is that the blind spot is, in
genuinely fuzzy tasks, unobservable: seeing it requires ground truth
independent of the verifier, and independent ground truth is exactly
what fuzzy tasks lack. Our method is a \emph{wind tunnel} --- a
controlled setting in which ground truth exists but the verifier is
denied it --- and the real-model measurements of Section~\ref{sec:real}
are what it produced.

\textbf{The wind tunnel.} Use tasks that carry a cheap, reliable oracle,
and \emph{blind the verifier to it}. On mathematical problem solving
\cite{hendrycks2021math,cobbe2021gsm8k}, the student emits a
reasoning trace and a final answer. The oracle is exact-match of the
answer against the gold solution --- cheap, reliable, and independent of
\(V\). The verifier is a frontier model asked whether the solution is
correct, given the problem and the trace but \emph{not} the gold answer.
This is a genuinely fuzzy verifier: it false-accepts wrong reasoning
that looks convincing and false-rejects correct reasoning that looks
unusual. Its blind spot is real, and --- because we hold the oracle in
reserve --- now measurable. The oracle is used for measurement only; it
never routes or trains, exactly as in a real deployment where it would
be absent.

This design pre-empts the two obvious objections. \emph{``Just use the
deterministic verifier''} is answered by including the oracle-verifier
as a control condition (\(\beta_{0} = 0\)): the point is to characterize
the fuzzy regime where no such verifier exists. \emph{``Math is not a
fuzzy task''} is answered by treating it as a \emph{model system}: math
carries a cheap gold oracle \emph{and} a frontier verifier that visibly
false-accepts and false-rejects, which is exactly what makes the blind
spot measurable here. Extending the wind tunnel to a genuinely fuzzy
task --- long-form claim verification or code-review quality, whose
oracle would have to be a \emph{super-verifier} (an expensive ensemble
validated against gold on the math tasks, then trusted where gold is
unavailable) --- is external-validity work we did not run;
Section~\ref{sec:limits} names it as the main open scope.

\textbf{Loop protocol.} Fix disjoint splits: a pool for round batches, a
held-out evaluation set used for all reported curves and never trained
on, and a fixed probe set for tracking error composition. Each round:
the student generates on a fresh batch; the verifier judges, blinded to
the oracle; rejects receive a fresh teacher correction; the student is
refit. To isolate \emph{data composition} as the independent variable
and remove optimizer path-dependence and catastrophic forgetting as
confounds, we retrain \emph{cumulatively from the base model} on all
data collected through round \(t\), rather than incrementally from
\(S_{t}\) (incremental training is retained only as a robustness check).
The verifier is pinned --- same model, prompt, temperature, and seed,
cached by output hash --- so that any change across rounds originates
from the student's inputs, never from drift in \(V\).

\textbf{Measurement and the decomposition that must not be skipped.}
Every round, on the held-out set and via the oracle, we record raw error
\(q_{t}\), escalation rate \(p_{\text{reject}}\), verifier recall
\(r_{t}\) and blind-spot rate \(\beta_{t}\), false-reject rate on
correct answers, verifier-estimated accuracy (the dashboard number), and
the user-facing error \(\varepsilon_{t}\). Critically,
\(\varepsilon_{t}\) is \emph{decomposed} into its accepted-and-wrong
term and its teacher-error term at all times; the conservation claim is
about the former alone, and conflating the two invites exactly the
critique a careful reviewer will raise. A sanity gate tracks whether
\(q_{t}\) actually falls: if the loop does not improve the student, the
conservation-\emph{floor} reading is unavailable --- an outcome we
report plainly (it is what the real-model runs of Section~\ref{sec:real}
in fact exhibit) rather than papering over, since the
dashboard-blindness result holds whether the student improves or
degrades.

This is the method; Section~\ref{sec:real} reports what it produced on
real language models, and Section~\ref{sec:exp} what it produced in a
controlled model where the loop provably improves the student.

\section{Real-model measurements}

\protect\phantomsection\label{sec:real}{}

This section is the paper's empirical core.\footnote{All code and data
  are public:
  \href{https://github.com/AltSlate-Labs/cascade-blindspot}{\texttt{github.com/AltSlate-Labs/cascade-blindspot}}.}
We ran the wind-tunnel method of Section~\ref{sec:design} on real
language models: students are Qwen2.5-Instruct (0.5B--32B), fine-tuned
with LoRA on a single H100; verifiers and teachers are OpenAI models
(gpt-4o-mini, gpt-4.1, gpt-5-mini), always blinded to the gold answer;
the oracle is exact-match on GSM8K \cite{cobbe2021gsm8k} and symbolic
equivalence on the hard subset (levels 4--5) of MATH
\cite{hendrycks2021math}. Nothing is assumed --- whether a blind spot
exists, how large it is, and how it moves with student and verifier
strength are all measured. Error bars throughout are \(95\%\) Wilson
intervals; the blind-spot rate \(\beta\) is a proportion over the
\emph{wrong-answer subset} only, so its intervals
(\(n \approx 70\)--\(170\)) are wider than those on the raw error rate
(\(n = 300\)). These are single-seed runs; Section~\ref{sec:limits}
treats seed variance.

\begin{figure}
\centering
\includegraphics[width=0.74\linewidth,height=\textheight,keepaspectratio]{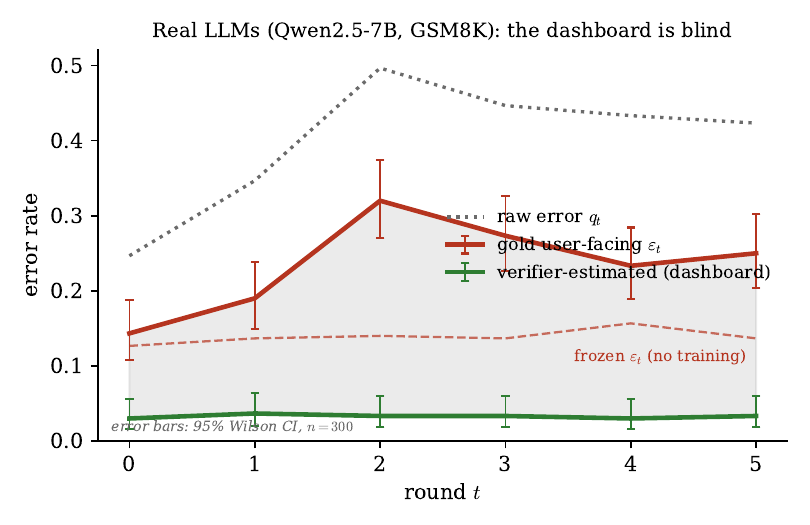}
\caption{Real LLMs (Qwen2.5-7B student, gpt-4o-mini verifier, GSM8K).
The dashboard (verifier-estimated error) stays \(\sim 3\%\) while gold
user-facing error climbs; the shaded gap is hidden harm. Error bars are
\(95\%\) Wilson intervals (\(n = 300\)); the gold--dashboard gap clears
them from round 1 on. The loop degrades the student here (raw error
rises) --- cross-family distillation, not improvement; the frozen
control (dashed) is flat.}
\label{fig:realsciss}
\end{figure}

\textbf{None of it shows on the dashboard (Figure~\ref{fig:realsciss}).}
Running the corrective loop with a Qwen2.5-7B student and a gpt-4o-mini
verifier on GSM8K, the verifier-estimated error of the delivered stream
holds flat near \(3\%\) across all rounds, while the true (gold)
user-facing error swings from \(14\%\) to \(32\%\) --- a
\(5\)--\(11 \times\) gap whose \(95\%\) intervals stop overlapping from
round 1 on, so no in-loop metric reveals it. The frozen control stays
near \(13\%\); the loop itself, trained on the frontier teacher's
corrections, \emph{degrades} the student (raw error rises), and the
dashboard is blind to the degradation exactly as it is blind to the
level. This is the alarming finding, and it does not depend on any
conjecture: the gold and dashboard curves are both directly measured.
This is a single training seed, so the exact per-round \emph{trajectory}
could shift; but the \emph{level} gap --- a flat \(\sim 3\%\) dashboard
against a \(14\)--\(32\%\) truth --- is far larger than any plausible
LoRA seed variance, so the result is the gap, not the particular curve
(Section~\ref{sec:limits}).

\begin{figure}
\centering
\includegraphics[width=0.66\linewidth,height=\textheight,keepaspectratio]{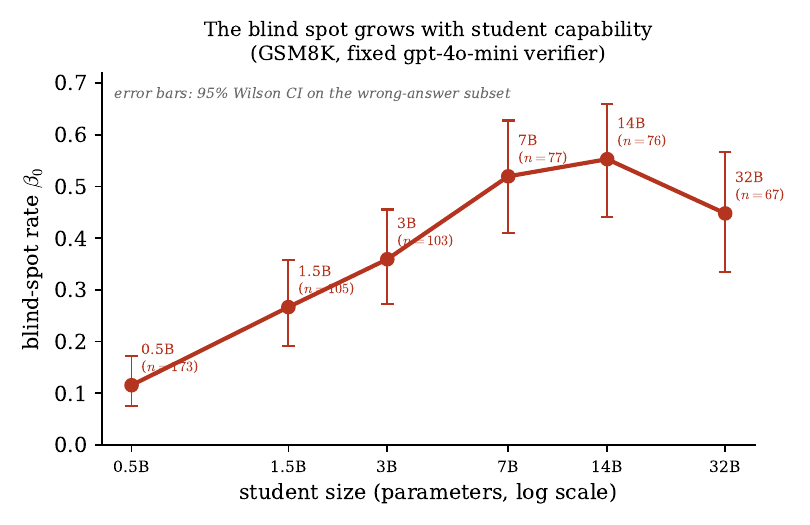}
\caption{Blind-spot rate \(\beta_{0}\) vs student size (GSM8K, fixed
gpt-4o-mini verifier; \(95\%\) Wilson bars over each size's wrong-answer
subset). A more capable student makes more \emph{convincing} wrong
answers, so the verifier accepts more of them; the rise saturates across
the largest three.}
\label{fig:betastu}
\end{figure}

\textbf{The blind spot grows with student capability
(Figure~\ref{fig:betastu}).} Holding the verifier fixed (gpt-4o-mini)
and sweeping the student 0.5B→32B on GSM8K, \(\beta_{0}\) rises from
\(0.12\) (CI \(\lbrack 0.08,0.17\rbrack\)) at 0.5B to \(0.55\)
(\(\lbrack 0.44,0.66\rbrack\)) at 14B: a stronger student makes subtler
errors, so the fixed verifier is fooled more often. The small-vs-large
separation sits well outside the intervals; the top three sizes
(7B--32B, all \(\lbrack \approx 0.35,0.65\rbrack\)) form a plateau
within noise, so the honest reading is a monotone rise that
\emph{saturates}, not a peak-and-fall. This is a \emph{different}
mechanism from Corollary 1 and worth separating from it. Corollary 1 is
about \emph{correlation} --- student and verifier finding the
\emph{same} wrong answer convincing because they share failure modes;
here the verifier is held fixed while only the student varies, so what
moves \(\beta_{0}\) is the \emph{capability gap}: a stronger student's
errors are intrinsically subtler and harder for \emph{any} fixed
verifier to catch, correlated or not. This capability-gap route to a
high blind spot is not part of the self-preference literature, and it
carries a sharper warning --- the blind spot gets \emph{worse} as
students improve, so the problem grows rather than shrinks with progress
on the cheap model. Note the finding is a \emph{static} characterization
--- a property of the student--verifier pair before any training --- so
it holds for a plain cascade and does not depend on the loop working;
that is exactly why it is robust.

\begin{figure}
\centering
\includegraphics[width=0.66\linewidth,height=\textheight,keepaspectratio]{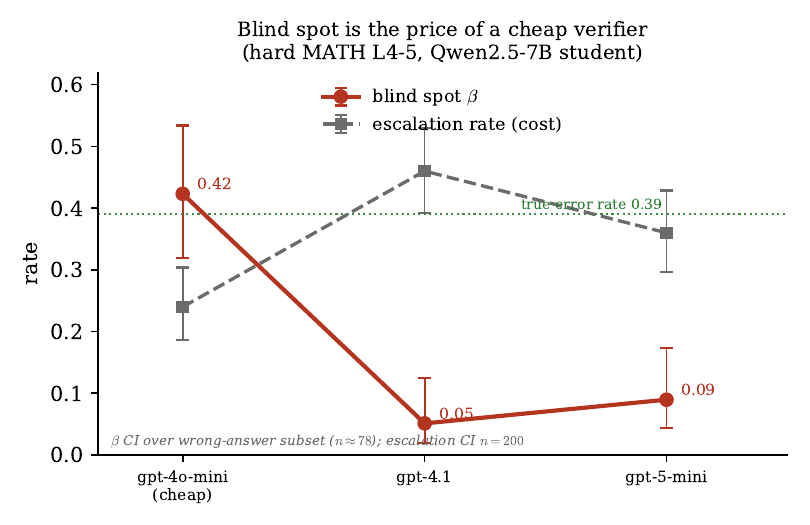}
\caption{Hard MATH (L4--5), fixed Qwen2.5-7B student (\(95\%\) Wilson
bars: \(\beta\) over the \(n \approx 78\) wrong answers, escalation over
\(n = 200\)). As the verifier strengthens, the blind spot \(\beta\)
falls but the escalation rate (cost) rises to meet the true error rate
(dotted) --- a strong verifier suppresses the blind spot by
over-escalating.}
\label{fig:betaver}
\end{figure}

\textbf{The blind spot is the price of a cheap verifier
(Figure~\ref{fig:betaver}).} On hard MATH, holding the 7B student fixed
and sweeping the verifier, \(\beta\) collapses from \(0.42\)
(gpt-4o-mini, CI \(\lbrack 0.32,0.53\rbrack\)) to \(0.05\)--\(0.09\) for
the two strong verifiers (gpt-4.1 and gpt-5-mini, intervals
\(\lbrack 0.02,0.12\rbrack\) and \(\lbrack 0.04,0.17\rbrack\) ---
statistically indistinguishable from each other): a strong verifier
catches almost every error on a task it can largely solve. But that low
blind spot is not free --- gpt-4.1's escalation rate (\(0.46\), CI
\(\lbrack 0.39,0.53\rbrack\)) \emph{meets or exceeds} the true error
rate (\(0.39\)), so it buys recall by over-rejecting correct answers
too, paying the frontier price on nearly half of queries --- the very
cost the cascade exists to avoid. The blind spot is thus largest
precisely in the \emph{cost-saving} configuration --- a capable cheap
student checked by a cheap verifier --- and buying it away gives the
cost back. Like the student sweep, this is a static operating-point
characterization, independent of any training.

\textbf{The loop degrades the student --- every teacher we tried.} The
one thing we set out to build, a loop that improves the small student,
we could not. The two figures above are inference-only sweeps; the
\emph{training} loop we ran on two students, the 1.5B and the 7B (GSM8K
corrective, and the teacher-family comparison of the next paragraph),
and on both it fails the same way. Across every teacher --- cross-family
(gpt-4o-mini and gpt-4o) and same-family (Qwen2.5-32B) --- naive LoRA
fine-tuning on the verifier-rejected tail \emph{degrades} the student
rather than improving it, and cumulatively \emph{collapses} it (raw
error rising to \(1\) as training on the hardest, most style-shifted
examples breaks its output format) --- an effect adjacent to the
model-collapse literature \cite{shumailov2024collapse}, and one the
dashboard is equally blind to. This is worth stating sharply: the
\emph{parametric} self-improving loop --- the weight-updating sibling of
today's training-free deferral-reuse methods \cite{intercascade2025,sarukkai2025deferral} --- fine-tunes on exactly the hard-tail
rejects most likely to destabilise a small student, so its naive form is
not merely suboptimal but actively harmful at this scale. So the real
runs establish the blind spot's \emph{structure} --- real, scaling up
with student and down with verifier, invisible to every in-loop metric
--- but \emph{not} the clean conservation \emph{scissors} (raw error
falling while user-facing error floors at \(q_{0}\beta_{0}\)), which
needs a loop that improves the student. The floor therefore stays a
\emph{theoretical} result --- derived in Section~\ref{sec:law} and
Appendix~\ref{sec:appendix}, validated synthetically in
Section~\ref{sec:exp} --- and its real-model demonstration (a stronger
student, or a training recipe that resists the hard-tail distribution
shift) is open. The next section explains why, whether the loop improves
or degrades the student, none of it registers on the dashboard.
Configuration and scripts are in the artifact (Section~\ref{sec:repro}).

\section{Why the dashboard is blind: a conservation-law account}

\protect\phantomsection\label{sec:law}{}

The measurements raise one question above the rest: why does \emph{no}
metric a practitioner can compute move, whether the loop is holding
steady or actively collapsing the student? The answer is structural, and
this section derives it. The same structure predicts that even a loop
that \emph{worked} --- that improved the student round over round ---
would leave a positive error floor rather than driving delivered error
to zero. We could not instantiate that improving loop at real-model
scale (Section~\ref{sec:real}), so the floor is a theoretical prediction;
the dashboard blindness it explains is not.

Consider the student's errors as two populations: those the verifier
detects (mass \(q_{t}r_{t}\)) and those it does not (mass
\(q_{t}\beta_{t}\)). The loop applies a correction pressure to the first
population and \emph{no direct pressure} to the second: the undetectable
errors receive no targeted training signal, though parameter updates
driven by the detectable ones can still spill into them
(Figure~\ref{fig:conserve}).

The user-facing error rate --- the quantity a deployment actually ships
--- is the error that survives verification:

\[\varepsilon_{t} = \underset{\text{ accepted \& wrong}}{\underbrace{q_{t}\beta_{t}}} + \underset{\text{ teacher error}}{\underbrace{\Pr\left\lbrack V\left( x,S_{t}(x) \right) = \text{ reject } \land O\left( x,G(x) \right) = \text{ wrong} \right\rbrack}}.\]

The second term is the teacher's fallibility on the rejected items; its
\emph{conditional} rate is bounded by \(G\)`s error rate, but its mass
shrinks with the escalation rate as the loop proceeds, and when \(G\)
and \(V\) share priors, wrong teacher labels that \(V\) would also
accept can enter training --- blurring the corrective/self-training
distinction (Section~\ref{sec:limits}). The first term is the blind spot,
and our central claim concerns it.

\textbf{Conjecture (Blind-spot conservation).} Under the corrective loop
with a fixed, imperfect verifier, and absent generalization spillover
into the blind region, the accepted-and-wrong error mass
\(q_{t}\beta_{t}\) is conserved across rounds even as the detectable
mass \(q_{t}r_{t}\) is driven down, so the user-facing error does not
vanish but asymptotes to
\[\varepsilon_{\infty} \approx q_{0}\beta_{0.}\] Spillover relaxes this
equality to an upper anchor,
\(\varepsilon_{\infty} \lesssim q_{0}\beta_{0}\), when \(\beta_{0}\) is
large relative to the student's capacity floor.
Appendix~\ref{sec:appendix} makes both statements precise (Proposition 1).

The intuition is that the loop removes error mass only where it has
signal. Detectable error \(q_{t}r_{t}\) shrinks because it is trained
against; undetectable error \(q_{t}\beta_{t}\) persists because it is
not. The anchor reaches the raw student too: with no direct signal on
the blind region, to first order \(q_{t}\) inherits the same floor, and
any drop below \(q_{0}\beta_{0}\) is spillover (self-healing). Only when
the loop trains on \emph{every} error --- the oracle-in-loop control C
of Section~\ref{sec:design} --- does raw error \(q_{t}\) fall to the
student's capacity limit (the dashed trajectory in
Figure~\ref{fig:conserve}) while delivered error \(\varepsilon_{t}\)
reaches zero. The synthetic study of Section~\ref{sec:exp} plots this
decomposition directly.

Two second-order effects perturb the anchor, and they push in
\emph{opposite} directions:

\begin{itemize}
\item
  \emph{Self-healing (corrective loop).} Gains in general competence may
  incidentally fix some blind-spot cases the loop never explicitly
  targeted, so the observed floor falls \emph{below} \(q_{0}\beta_{0}\).
  The size of this gap measures how much reliability the loop delivers
  ``for free'' beyond what the verifier can see.
\item
  \emph{Self-reinforcement (self-training loop).} When accepted student
  outputs are fed back as positive targets, false accepts are injected
  as \emph{labels}, actively teaching the student to reproduce the
  errors the verifier cannot catch. The floor then sits strictly
  \emph{above} the corrective floor, and --- when reinforcement
  outweighs self-healing --- above \(q_{0}\beta_{0}\) itself, with
  user-facing quality able to degrade while every in-loop signal
  improves.
\end{itemize}

The falsifiable core is narrower than this interpretive frame and should
not be confused with it. What can be \emph{refuted} is H1 --- a strictly
positive floor, against the null \(\varepsilon_{\infty} = 0\) --- and H2
--- the floor increasing in the \emph{measured} \(\beta_{0}\), against
the null of no relation. The anchor \(q_{0}\beta_{0}\) is a reference
scale for the corrective loop, not a conserved quantity; a floor of
zero, or a floor uncorrelated with \(\beta_{0}\), would count against
the framework. The self-healing / self-reinforcement split then
interprets \emph{where} a given loop lands relative to the anchor.

Two corollaries sharpen the practical stakes.

\textbf{Corollary 1 (Correlation scaling).} The floor is increasing in
the verifier's blind-spot rate \(\beta_{0}\), and \(\beta_{0}\) is large
when student and verifier share failure modes --- when a wrong answer
convincing to \(S\) is also convincing to \(V\). (Shared failure modes
are one sufficient cause of a high \(\beta_{0}\), not the only one --- a
lazy rubber-stamping verifier has high \(\beta_{0}\) with no shared
bias, and, as Section~\ref{sec:real} measures directly, so does a large
\emph{capability gap}: a strong student's errors are intrinsically
subtle, so even an \emph{uncorrelated} verifier catches fewer of them.
Correlation and capability-gap are distinct routes to the same high
\(\beta_{0}\).) Verifiers from the same family or pretraining lineage as
the student should therefore yield higher floors than independent
verifiers; a deterministic oracle-verifier (\(\beta_{0} = 0\)) yields no
floor. The engineering reading is a design rule: \emph{decorrelate the
verifier from the student} --- and, from the capability-gap route,
expect the blind spot to \emph{grow} as the cheap student improves.

\textbf{Corollary 2 (Dashboard blindness).} Every metric a practitioner
naturally monitors --- the escalation rate \(p_{\text{reject}}\) (a
proxy for cost) and the verifier's estimated accuracy on the accepted
stream --- is computed \emph{through} \(V\). As the loop concentrates
error into \(V\)`s blind spot, these metrics improve or hold steady ---
fewer escalations, near-perfect apparent accept-stream quality
(Figure~\ref{fig:dash}), the accepted-stream component being structural
since \(V\) cannot flag its own accepts. The gold-truth accuracy of the
accepted stream, which no in-loop instrument reports, fails to improve
in step --- it stays floored. The system is thus blind to its own
degradation \emph{by construction}, not by oversight; the cost dashboard
and the quality dashboard are the same instrument, and it is the
compromised one.

\textbf{The account's five predictions.} The conservation-law account
makes five directional, falsifiable predictions, tested across the rest
of the paper. One is already confirmed on real models: the
dashboard-blindness prediction (H4) is directly measured
(Section~\ref{sec:real}). A second is addressed only at the level of its
\emph{precondition}: H2 asks whether the error \emph{floor} scales with
\(\beta_{0}\), and while we never measure a floor on real models, we do
measure that \(\beta_{0}\) itself rises with the student--verifier
capability gap (Section~\ref{sec:real}) --- the necessary precursor, not
the floor-scaling itself. The floor (H1), the loop-design hazard (H3),
and the mitigation exchange rates (H5) all need a loop that improves the
student, so they are tested in the controlled model of
Section~\ref{sec:exp}. Each is stated with its null.

\begin{description}
\item[H1 (Positive floor)]
User-facing error \(\varepsilon_{t}\) does not vanish but asymptotes to
a strictly positive floor \(\lesssim q_{0}\beta_{0}\); to first order
raw error \(q_{t}\) inherits the same floor, and a matched
\(\beta_{0} = 0\) control isolates the verifier-induced excess, while
the oracle-in-loop control drives delivered error to zero. \emph{Null}:
\(\varepsilon_{t} \rightarrow 0\) --- no floor.
\item[H2 (Blind-spot scaling)]
The asymptotic floor \(\varepsilon_{\infty}\) is increasing in the
measured initial blind-spot rate \(\beta_{0}\). Sweeping the verifier to
vary \(\beta_{0}\) traces a positive relation; an oracle-verifier sits
at the origin. \emph{Null}: the floor is independent of \(\beta_{0}\).
\item[H3 (Loop-design hazard)]
The self-training loop yields a strictly higher floor than the
corrective loop; under strong enough reinforcement it can push the floor
above \(q_{0}\beta_{0}\) and render \(\varepsilon_{t}\) non-monotone
while in-loop metrics improve. \emph{Null}: the two loops floor at the
same level.
\item[H4 (Dashboard blindness)]
A large, persistent gap separates the verifier-estimated error of the
delivered stream from its gold error, invisible to every in-loop metric.
A further prediction, untested here, is that the gap \emph{widens} over
rounds whenever self-training grows the blind mass. \emph{Null}:
estimated and gold error track each other.
\item[H5 (Mitigation exchange rate)]
A decorrelated verifier ensemble, and spending a fixed fraction of
budget on random oracle audits that re-inject blind-spot cases into
training, each lower the floor toward the oracle-in-loop control, at a
quantifiable cost. \emph{Null}: neither intervention moves the floor.
\end{description}

\section{Synthetic validation}

\protect\phantomsection\label{sec:exp}{}

Because the real-model loop degrades rather than improves the student
(Section~\ref{sec:real}), the conservation \emph{floor} --- the behaviour
of a loop that \emph{works} --- cannot be read off the real runs. We
therefore turn to a controlled model in which the loop provably improves
the student, and ask whether the law appears when the blind spot's
dynamics are \emph{emergent} rather than imposed. A positive answer is
necessary, not sufficient --- it cannot speak to real language models
--- but a negative answer would refute the mechanism outright. This is
where H1, H3, and H5 are tested.

\textbf{Model.} The task is \(K = 10\)-way classification standing in
for ``produce the right answer''; ground truth \(y^{\ast (\varphi)}\) is
a fixed random two-layer network. The \emph{student} is a logistic model
on random features, retrained each round on a pool the loop grows; it
improves with data. The \emph{verifier} is fixed with two regimes: on a
\emph{blind region} --- a fixed random half-space whose threshold is set
so the region covers input-space mass \(\rho\) --- it rubber-stamps
whatever the student says; elsewhere it re-derives its own class with a
strong classifier and accepts only on agreement. So \(\rho\) sets the
verifier's blind-spot rate; the measured \(\beta_{0}\) slightly exceeds
\(\rho\) (e.g. \(0.76\) at \(\rho = 0.7\)) because the strong classifier
occasionally agrees with a wrong answer outside the blind region. This
models blind \emph{mass}, not correlation per se --- H2 tests
floor-against-\(\beta_{0}\), and Corollary 1's correlation reading is
one account of what makes \(\beta_{0}\) large. Crucially, nothing tells
the loop to spare blind-spot errors: blind-region items are accepted, so
they never enter the training pool; whether that yields a conserved
floor, self-healing, or self-reinforcement is left to the learning
dynamics. We run five variants --- corrective (A), self-training (B),
oracle-in-loop (C, perfect verifier trained on all items), frozen (D),
and a matched control (perfect verifier, trained on wrong items only,
isolating the verifier-induced excess) --- for eight rounds over 20
seeds, the teacher supplying true labels on rejected items. The
dashboard metric is the verifier-estimated error of the delivered stream
--- the fraction the pinned verifier rejects on a re-pass --- which a
practitioner computes without gold.

\begin{longtable}[]{@{}lcccccc@{}}
\caption{Per-variant outcomes at \(\rho = 0.7\) (20 seeds, mean \(\pm\)
sd). Frozen sits \emph{on} the anchor \(q_{0}\beta_{0} = 0.334\);
\emph{both} learning variants land below it (self-training above
corrective); oracle reaches zero. The verifier-induced excess in raw
error is
\(q_{T}^{\text{corr }} - q_{T}^{\text{matched }} = .304 - .283 = .021\).}\tabularnewline
\toprule\noalign{}
variant & \(q_{0}\) & \(\beta_{0}\) & \(q_{T}\) & \(\varepsilon_{T}\) &
\(q_{0}\beta_{0}\) & \(\text{dash}_{T}\) \\
\midrule\noalign{}
\endfirsthead
\toprule\noalign{}
variant & \(q_{0}\) & \(\beta_{0}\) & \(q_{T}\) & \(\varepsilon_{T}\) &
\(q_{0}\beta_{0}\) & \(\text{dash}_{T}\) \\
\midrule\noalign{}
\endhead
\bottomrule\noalign{}
\endlastfoot
corrective (A) & .438 & .764 & \(.304 \pm .01\) &
\(\mathbf{.249 \pm .01}\) & .334 & .046 \\
self-training (B) & .438 & .764 & \(.356 \pm .02\) &
\(\mathbf{.296 \pm .01}\) & .334 & .045 \\
oracle-in-loop (C) & .438 & .000 & \(.248 \pm .01\) & \(\mathbf{.000}\)
& --- & .000 \\
\(\beta_{0} = 0\) matched & .438 & .000 & \(.283 \pm .01\) & \(.000\) &
--- & .000 \\
frozen (D) & .438 & .764 & \(.438 \pm .02\) & \(\mathbf{.334 \pm .02}\)
& .334 & .048 \\
\label{tab:variants}
\end{longtable}

\begin{figure}
\centering
\includegraphics[width=0.72\linewidth,height=\textheight,keepaspectratio]{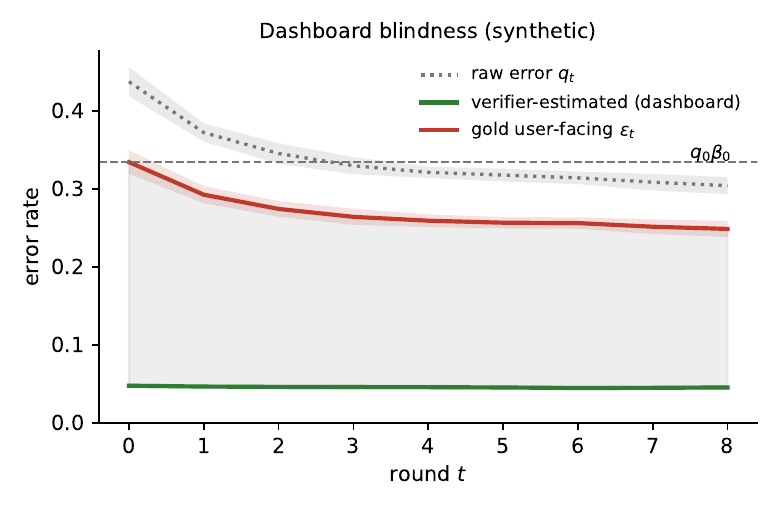}
\caption{\emph{Dashboard blindness} (\(\rho = 0.7\), \(\pm 1\) sd
bands). The verifier-estimated error of the delivered stream holds near
\(4.6\%\) while the gold error falls and floors at \(\approx 24.9\%\);
the shaded gap is a large, persistent hidden harm no verifier-computed
dashboard reports. Raw error \(q_{t}\) (dotted) also floors, near
\(q_{0}\beta_{0}\).}
\label{fig:dash}
\end{figure}

\begin{figure}
\centering
\includegraphics[width=0.72\linewidth,height=\textheight,keepaspectratio]{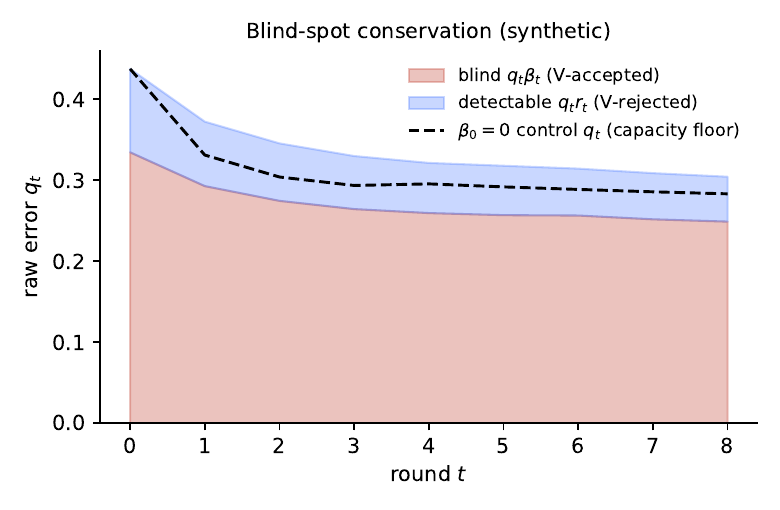}
\caption{\emph{Conservation} (\(\rho = 0.7\)). Raw error \(q_{t}\)
splits into a detectable band the loop trains away and a blind band it
receives no direct signal on; the blind mass is largely conserved. The
dashed line is the matched \(\beta_{0} = 0\) control --- the capacity
floor the same student reaches when it \emph{can} train on the
blind-region errors --- so the gap to it is the verifier-induced
excess.}
\label{fig:conserve}
\end{figure}

\begin{figure}[t]
\centering
\begin{minipage}[t]{0.49\linewidth}\centering
\includegraphics[width=\linewidth]{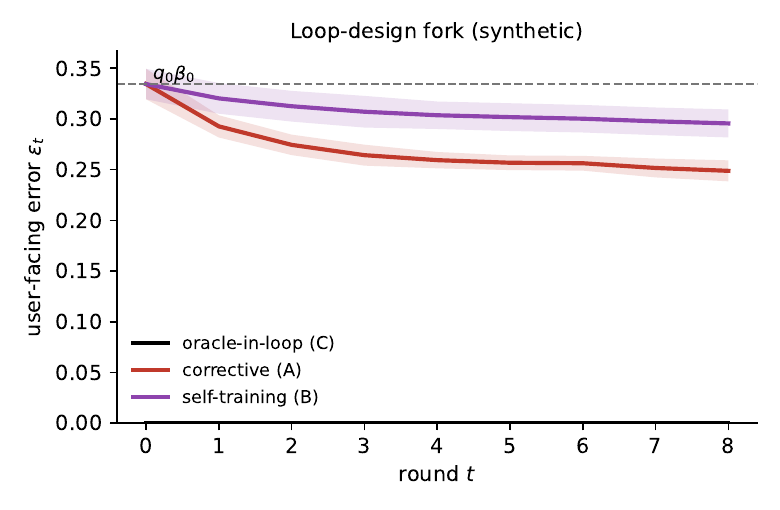}\\[2pt]{\small (a) loop-design fork}
\end{minipage}\hfill
\begin{minipage}[t]{0.49\linewidth}\centering
\includegraphics[width=\linewidth]{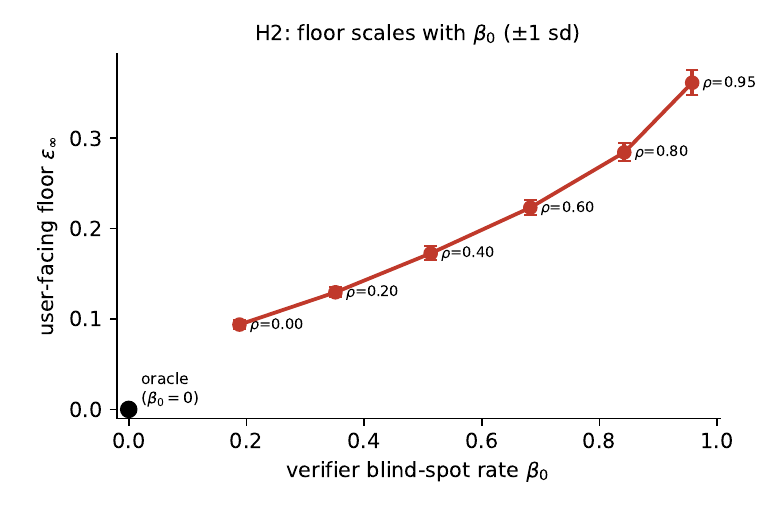}\\[2pt]{\small (b) floor vs $\beta_0$ (H2)}
\end{minipage}
\caption{(a) User-facing error by loop variant: oracle-in-loop (C) reaches zero, and both learning loops land below $q_0\beta_0$---corrective (A) lowest, self-training (B) above corrective but still below the anchor. (b) The user-facing floor scales monotonically with the verifier's blind-spot rate $\beta_0$ ($\pm 1$ sd), with the oracle at the origin.}
\label{fig:fork}
\end{figure}

\textbf{Results.} Four of the five predictions are supported as stated,
and H3 is supported in a corrected form (Table~\ref{tab:variants}).
\emph{H1 (positive floor)}: under the fuzzy verifier, gold user-facing
error falls but floors --- from \(\varepsilon_{0} = 0.33\) to
\(\varepsilon_{\infty} = 0.25 \pm .01\) --- while the oracle-in-loop
control reaches \(0\); a matched \(\beta_{0} = 0\) control (perfect
verifier, same reject-and-retrain rule) isolates the verifier's effect
on raw error,
\(q_{T}^{\text{corr }} - q_{T}^{\text{matched }} = 0.304 - 0.283 = 0.021\)
--- this excess conflates two verifier-caused effects, never labelling
the blind region and the smaller training pool that results, so it is
verifier-induced but not blind-mass censoring alone
(Figure~\ref{fig:conserve}). \emph{H4 (dashboard blindness)}: the
dashboard reads \(4.6\%\) while the truth is \(24.9\%\) --- a large
\emph{level} gap. It does not widen here: under both loops gold error
falls while the dashboard stays flat, so the gap narrows; the widening
form needs the blind mass to grow, which our readily-generalising
student does not exhibit, so it remains a real-model prediction
(Figure~\ref{fig:dash}). \emph{H2 (scaling)}: the floor increases
monotonically with the measured \(\beta_{0}\), from \(0.09\) to \(0.36\)
as \(\beta_{0}\) sweeps \(0.19 \rightarrow 0.96\), with tight per-seed
bands (Figure~\ref{fig:fork}b); at the smallest \(\beta_{0}\) the floor
slightly exceeds \(q_{0}\beta_{0}\) because the capacity floor binds, so
the \(\lesssim q_{0}\beta_{0}\) reading holds for \(\beta_{0}\) large
relative to that capacity limit. \emph{H3 (loop-design hazard)}:
self-training floors strictly higher than corrective
(\(\varepsilon_{T} = 0.30\) vs \(0.25\); \(q_{T} = 0.36\) vs \(0.30\));
its raw error \emph{falls} over rounds (from \(0.438\) to \(0.356\)),
not rises --- the predicted absolute rise did not occur in this
instantiation, so the effect here is reinforcement \emph{relative to}
the corrective loop, and the non-monotone regime remains a real-model
prediction (Figure~\ref{fig:fork}a).

\begin{figure}
\centering
\includegraphics[width=0.76\linewidth,height=\textheight,keepaspectratio]{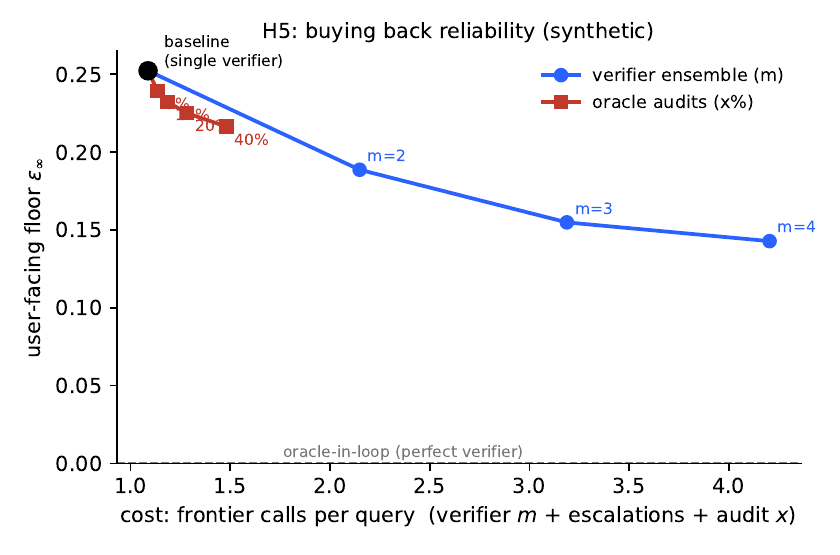}
\caption{\emph{Mitigation exchange rate} (\(\rho = 0.7\)). A
cost--reliability Pareto: a verifier ensemble (blue) buys large
reductions in the floor at steep cost (\(m\) verifier calls per query);
random oracle audits (red) are cheap but shallow. The oracle-in-loop
bound (dashed) is a perfect verifier.}
\label{fig:mit}
\end{figure}

\textbf{H5 (mitigation exchange rate).} Both proposed remedies move the
floor toward the oracle bound, and they occupy different regions of the
cost--reliability plane (Figure~\ref{fig:mit}). A decorrelated verifier
ensemble --- \(m\) verifiers with independent blind regions, rejecting
on any dissent --- drives the effective blind-spot rate toward
\(\rho^{m}\) plus a residual agreement term (measured \(\beta_{0}\):
\(0.76 \rightarrow 0.40\) at \(m = 4\), above \(\rho^{4} \approx 0.24\)
because the strong classifier still false-accepts occasionally outside
the blind region) and cuts the floor from \(0.25\) to \(0.14\), but pays
\(m\) verifier calls per query. Random oracle audits are far cheaper yet
weaker: auditing \(40\%\) of items lowers the floor only to \(0.22\),
because most of the audit budget lands on non-blind items --- an
untargeted audit spends most of its checks where the verifier already
sees. Neither reaches the oracle floor within the tested budget; the
ensemble is the stronger lever, and the obvious refinement --- auditing
where the verifier is least certain rather than at random --- is left to
the real-model study.

\textbf{The one honest correction.} The conjecture anchored the
corrective floor at \(q_{0}\beta_{0} = 0.33\). Both learning variants
land \emph{below} it --- corrective at \(0.25\), self-training at
\(0.30\) --- and only the frozen control sits \emph{on} the anchor
(\(0.33\), with no learning to spill). The predicted \emph{above}-anchor
regime for self-training was \emph{not} observed: self-healing dominates
even when accepted outputs are recycled as labels, so self-training
floors above corrective rather than above \(q_{0}\beta_{0}\). The
ordering that held is frozen \(\left( = q_{0}\beta_{0} \right)\)
\textgreater{} self-training \textgreater{} corrective \textgreater{}
oracle, which still separates the mechanisms, and the equality should be
read as \(\varepsilon_{\infty} \lesssim q_{0}\beta_{0}\) for the
corrective loop, the \(0.33 - 0.25\) gap quantifying self-healing.
Pushing self-training across the anchor would need reinforcement strong
enough to outweigh this spillover --- a plausible real-model regime our
synthetic student, which generalises readily, does not reach.

\textbf{What this does and does not establish.} Two confirmations are
close to structural: because blind-region items are censored from the
training pool, a persistent floor (H1) and its growth with blind mass
(H2) follow almost arithmetically --- they show the mechanism is
\emph{self-consistent}, not that it is large in practice. The genuinely
informative outcomes are those the censoring does not force: the size of
the self-healing gap, the H3 ordering (and the \emph{absence} of an
absolute rise), the \(0.021\) verifier-induced excess against the
matched control, and the H5 exchange rates. In all cases the blind
spot's \emph{existence} is modelled (through \(\rho\)) while only its
\emph{dynamics under the loop} are emergent; none of it evidences the
effect's magnitude on real language models --- which is why the
real-model measurements of Section~\ref{sec:real}, not this study, carry
the paper's empirical weight. What the synthetic model adds is the one
thing the real runs could not supply: the behaviour of a loop that
\emph{improves} the student, and thus direct evidence for the
conservation floor the theory predicts.

\section{Threats to validity}

The measurements and the synthetic study are each exposed to
characteristic failures, and we address them in turn. The floor claim
depends on the loop improving the student; on real models it did not,
and rather than treat that as a void premise we report it as a finding
(Section~\ref{sec:real}) and fall back to the controlled model, where the
sanity gate on \(q_{t}\) confirms the loop does improve. In that
controlled study, so that forgetting or optimizer instability cannot
masquerade as decoupling, we train cumulatively from the base model and
include the frozen-student control (D) to separate the two. If the
blind-spot mass self-heals, that is not a failure but the negative
result the conjecture already anticipates, named by the sign of the
deviation from \(q_{0}\beta_{0}\). The model-system critique --- that
math is not a genuinely fuzzy task --- is met partly by the real-model
measurements of Section~\ref{sec:real}, where the frontier verifier
visibly false-accepts and false-rejects on GSM8K and hard MATH;
extending to a fully fuzzy task via a validated super-verifier is named
as open scope (Section~\ref{sec:limits}), not claimed. Verification cost
--- the dominant expense, since \(V\) is a frontier model --- is
controlled by deterministic caching of the pinned verifier.

The correlation corollary is the claim most exposed to challenge,
precisely because it borders the self-preference literature
\cite{panickssery2024selfpreference}. The defense is built into the
measurement: H2 is stated over the \emph{measured} \(\beta_{0}\), with
model family used only as a manipulation to spread \(\beta_{0}\) across
a range. The relation between the floor and \(\beta_{0}\) therefore
stands whether or not self-preference is the reason \(\beta_{0}\) is
high for same-family pairs.

\section{Limitations}

\protect\phantomsection\label{sec:limits}{}

Distinct from the threats above --- which defend the proposed design ---
these bound the paper's own claims. (1) The synthetic blind region is a
\emph{static} input-space set, whereas real verifier blind spots are
output-dependent and move with the student's error distribution
(assumption A1 of Appendix~\ref{sec:appendix}); a soft or moving blind
region could weaken conservation. (2) The anchor \(q_{0}\beta_{0}\)
assumes a \emph{stationary} query distribution. (3) When the teacher
\(G\) is the verifier \(V\), the teacher-error term is not independent
of \(\beta\) and the floor formula is optimistic --- wrong labels \(V\)
would also accept enter training. (4) A \(10\)-way classification task
with a logistic student may not transfer to open-ended generation, where
``the same wrong answer'' is itself ill-defined. (5) The H5 exchange
rates are properties of the synthetic geometry, not portable constants.
(6) The real-model runs are \emph{single-seed}: point estimates carry
\(95\%\) Wilson intervals (Section~\ref{sec:real}), but seed-to-seed
variance in LoRA training is not bounded, so the per-round
\emph{trajectory} of the degrading loop should be read as one
representative run, not an averaged curve --- the static blind-spot
sweeps, being training-free, do not share this caveat. The real-model
measurements (Section~\ref{sec:real}) and the appendix model each resolve
a subset of these.

\section{Implications}

Three consequences follow for anyone building cost-saving cascades with
verifier feedback; the first two rest on what we \emph{measured}, the
third on the theory. First --- measured --- \emph{the cost dashboard
lies}: escalation rate and accept-stream accuracy read healthy (a flat
\(3\%\)) while true delivered error swings to \(32\%\), because both are
computed through the verifier and improve or hold steady as error hides
in its blind spot. An independent audit channel --- a small, periodic
gold-labeled sample --- is not optional instrumentation but the only
instrument that can see the effect. Second --- measured ---
\emph{verifier choice is a reliability decision, not only a cost
decision}: the blind spot grows with student capability and shrinks with
verifier capability, so a cheaper verifier trades reliability for cost
on an axis no dashboard shows, and buying the blind spot away with a
frontier verifier returns the cost saving it was meant to provide. Third
--- from the theory --- \emph{the loop-design choice between corrective
and self-training is a safety choice}: feeding accepted outputs back as
positive labels converts a passive blind spot into an actively
reinforced one.

None of this argues against cascades, which remain the most direct cost
lever available, nor against closing the loop, which genuinely lowers
cost. It argues that the reliability of a self-improving cascade must be
measured \emph{outside} the verifier that defines it, because a system
optimized against a proxy will, given the chance, satisfy the proxy
rather than the goal --- and a cascade that retrains on its own verifier
is given exactly that chance, round after round.

\section{Conclusion}

We measured the blind spot of a cost-saving cascade on real LLMs and
found it moves adversarially: it grows with student capability and
shrinks with verifier capability, so it is largest exactly in the
cheap-student, cheap-verifier regime that makes cascades attractive, and
buying it away with a frontier verifier returns the cost saving by
escalating on nearly half of queries. We found that closing the loop
with naive corrective fine-tuning does not improve a small student but
degrades and collapses it, across every teacher --- a direct caution
against the parametric self-improving loop that fine-tunes the student
on the verifier's rejects. And we found that none of this registers on
any metric a practitioner can compute: the dashboard reads a flat
\(3\%\) while true delivered error swings to \(32\%\). To explain that
blindness we gave a two-population \emph{conservation law} --- the loop
trains only on verifier-detectable errors, so user-facing error
asymptotes to a floor \(\lesssim q_{0}\beta_{0}\) rather than vanishing,
and every in-loop metric improves while true quality does not ---
derived in a linear model (Appendix~\ref{sec:appendix}) and validated in a
synthetic study where the loop provably improves the student
(Section~\ref{sec:exp}). The clean floor itself we leave as theory,
because no real-model loop we ran reached it. The through-line is
practical: the reliability of a self-improving cascade cannot be read
from any metric computed through its own verifier --- it must be
measured outside.

\appendix
\section{Two-population model of the error floor}

\protect\phantomsection\label{sec:appendix}{}

Track the two error masses \(d_{t} = q_{t}r_{t}\) (detectable) and
\(b_{t} = q_{t}\beta_{t}\) (blind), with raw error
\(q_{t} = d_{t} + b_{t}\) and delivered error
\(\varepsilon_{t} = b_{t} + \tau_{t}\), where \(\tau_{t}\) is the
teacher's error on rejected items. We model one round of each loop as a
linear update under idealised assumptions: \emph{(A1)} the verifier's
blind region is fixed; \emph{(A2)} the query distribution is stationary;
\emph{(A3)} each round the loop corrects a fraction
\(c \in (0,1\rbrack\) of the detectable mass, transfers a fraction
\(\gamma \geq 0\) of that reduction to the blind mass by generalisation
(spillover), and --- under self-training --- recycles false accepts as
labels, adding back \(\alpha \geq 0\) of the blind mass:

\[d_{t + 1} = (1 - c)d_{t},\quad b_{t + 1} = b_{t} - \gamma\left( d_{t} - d_{t + 1} \right) + \alpha b_{t}.\]

\textbf{Proposition 1.} Under (A1)--(A3), with
\(\tau_{t} \rightarrow 0\):

\begin{enumerate}
\item
  \emph{(Frozen, \(c = 0\), \(\alpha = 0\).)} \(b_{t} \equiv b_{0}\), so
  \(\varepsilon_{\infty} = q_{0}\beta_{0}\): exact conservation.
\item
  \emph{(Corrective, \(c > 0\), \(\alpha = 0\).)}
  \(d_{t} \rightarrow 0\) and
  \[\varepsilon_{\infty} = q_{0}\beta_{0} - \gamma d_{0} \leq q_{0}\beta_{0}\quad\left( \gamma d_{0} \leq b_{0} \right),\]
  with equality iff \(\gamma = 0\); the slack \(\gamma d_{0}\) is
  self-healing.
\item
  \emph{(Self-training, \(\alpha > 0\).)} at every finite horizon \(T\),
  \[b_{T} = b_{0} - \gamma\left( d_{0} - d_{T} \right) + \alpha\sum_{t < T}b_{t} > b_{\infty}^{\text{corr}},\]
  strictly once \(\alpha > 0\), and \(b_{T} > q_{0}\beta_{0}\) once the
  accumulated reinforcement outweighs \(\gamma d_{0}\). The linear
  reinforcement has no finite limit (blind mass would grow without bound
  as \(d_{t} \rightarrow 0\)), so this is a finite-horizon statement; a
  realistic \emph{saturating} reinforcement --- \(\alpha\) acting only
  on the not-yet-reinforced mass, capping total error at \(1\) --- gives
  a genuine elevated fixed point.
\end{enumerate}

\emph{Proof sketch.} For (i)--(ii),
\(d_{t} = (1 - c)^{t}d_{0} \rightarrow 0\) when \(c > 0\) (and
\(d_{t} \equiv d_{0}\) when \(c = 0\)); since \(\varepsilon = b + \tau\)
with \(\tau \rightarrow 0\) depends on the blind mass \(b\), not on
\(d\), the frozen case gives
\(\varepsilon_{\infty} = b_{0} = q_{0}\beta_{0}\) even though \(d\)
never shrinks --- persistent detectable mass only keeps escalation cost
high. For (ii) the detectable reductions telescope,
\(\sum_{t}\left( d_{t} - d_{t + 1} \right) = d_{0} - d_{\infty} = d_{0}\),
so \(b_{\infty} = b_{0} - \gamma d_{0}\). In (iii) each step adds
\(\alpha b_{t} > 0\), so \(b_{T}\) strictly exceeds the corrective floor
at every \(T\); the linear term diverges, hence the finite-horizon
phrasing.

The synthetic study (Section~\ref{sec:exp}) instantiates this with
\(q_{0}\beta_{0} = 0.334\),
\(d_{0} = q_{0}\left( 1 - \beta_{0} \right) = 0.103\), and corrective
\(\varepsilon_{\infty} = 0.249\), giving a fitted spillover
\(\gamma = (0.334 - 0.249)/0.103 \approx 0.83\): generalisation
transfers most of the detectable correction into the blind region, which
is why the corrective floor sits well below the anchor. Self-training's
delivered error at \(T = 8\), \(\varepsilon_{T} = 0.296\), lands between
the corrective floor and \(q_{0}\beta_{0}\) --- over this horizon its
reinforcement offsets, but does not overcome, the spillover. Assumption
(A1), the static blind region, is the one Section~\ref{sec:limits} flags
as least realistic; a moving blind region is precisely what a real-model
study with output-dependent verifiers would probe.

\section{Reproducibility}

\protect\phantomsection\label{sec:repro}{}

All code, data, and figure scripts are public at
\href{https://github.com/AltSlate-Labs/cascade-blindspot}{\texttt{github.com/AltSlate-Labs/cascade-blindspot}}.
The committed measurement files regenerate every figure without a GPU or
API key; re-running the measurements from scratch needs one GPU and an
OpenAI key.

All synthetic results regenerate from two self-contained scripts ---
\texttt{phase0\_synthetic.py} (H1--H4) and \texttt{h5\_mitigation.py}
(H5) --- in a few minutes of CPU each, at fixed seeds; the five figures
are their direct output. Configuration: input dimension \(20\),
\(K = 10\) classes, target a fixed random two-layer network (64 hidden,
\(\tanh\)); student = multinomial logistic regression on \(300\) random
ReLU features (\(C = 3\)); verifier = the same on \(450\) features fit
on 5000 labelled points (\(C = 6\)), with a blind region set at the
\(\rho\)-quantile of a fixed random projection; cumulative-from-base
training with base pool \(140\), batch \(600\)/round, held-out eval
3000, \(8\) rounds, \(20\) seeds; the reported floor is the
last-three-round mean of \(\varepsilon\). The ensemble of
Figure~\ref{fig:mit} uses \(m\) independent blind half-spaces. numpy
2.3, scikit-learn 1.8.

The real-model results (Section~\ref{sec:real}) regenerate from
\texttt{run\_phase0.py} (the loop), \texttt{sweep\_capability.py}
(student sweep), and \texttt{math\_sweep.py} (verifier sweep on hard
MATH), with \texttt{make\_real\_figures.py} rendering the figures.
Students are Qwen2.5-Instruct (0.5B--32B) via transformers + LoRA (rank
16 on all seven linear modules, lr \(10^{- 5}\) with warmup) on one
H100; verifiers and teachers are the OpenAI API (gpt-4o-mini, gpt-4.1,
gpt-5-mini), called concurrently and blinded to the gold answer; the
oracle is GSM8K exact-match and \texttt{math-verify} symbolic
equivalence on MATH-500 levels 4--5; \(300\) held-out problems per
condition. torch 2.13, transformers, peft.

\bibliographystyle{unsrtnat}
\bibliography{refs}
\end{document}